\documentclass[runningheads]{llncs}
\usepackage{graphicx}

\usepackage{tikz}
\usepackage{comment}
\usepackage{amsmath,amssymb} 
\usepackage{color}
\usepackage{bmpsize}

\usepackage{algorithmic}
\usepackage{textcomp}
\usepackage{xcolor}
\usepackage{acronym}
\usepackage{lscape}
\usepackage{amsmath}
\usepackage{textcomp}
\usepackage{amssymb}
\usepackage{comment}
\usepackage{epsfig}
\usepackage{epstopdf}

\newcommand*{\ie}		{i.e.,\ }

\acrodef{AI} 						{Artificial Intelligence}
\acrodef{ML} 						{Machine Learning}
\acrodef{DL} 						{Deep Learning}
\acrodef{SVM} 						{Support Vector Machine}
\acrodef{ANN} 						{Artificial Neural network}
\acrodef{CNNs} 						{Convolutional Neural Networks}
\acrodef{CNN} 						{Convolutional Neural Network}

\acrodef{3DPM} 						{3-Dimensional Particle Measurement}

\begin{document}
\pagestyle{headings}
\mainmatter
\def\ECCVSubNumber{-}  

\title{Depth Contrast: Self-Supervised Pretraining on 3DPM Images for Mining Material Classification} 


\titlerunning{Depth Contrast: Self-Supervised Pretraining on 3DPM Images for Mining Material Classification}
%
\author{Prakash Chandra Chhipa\inst{1,*} \and
Richa Upadhyay\inst{1} \and
Rajkumar Saini\inst{1} \and
Lars Lindqvist\inst{2} \and
Richard Nordenskjold\inst{2} \and
Seiichi Uchida\inst{3} \and
Marcus Liwicki\inst{1}}
\authorrunning{P.C. Chhipa et al.}
%
\institute{Machine Learning Group, EISLAB, Lule\aa~Tekniska Universitet, Lule\aa, Sweden \email{first.middle.last@ltu.se} \and
Optimation Advanced Measurements AB, Lule\aa, Sweden \email{first.last@optimation.se} \and
Human Interface Laboratory, Kyushu University, Fukuoka, Japan \email{uchida@ait.kyushu-u.ac.jp} \\
\textit{\textsuperscript{*}Corresponding author - prakash.chandra.chhipa@ltu.se}
}
\maketitle

\begin{abstract}
This work presents a novel self-supervised representation learning method to learn efficient representations without labels on images from a 3DPM sensor (3-Dimensional Particle Measurement; estimates the particle size distribution of material) utilizing RGB images and depth maps of mining material on the conveyor belt. Human annotations for material categories on sensor-generated data are scarce and cost-intensive. Currently, representation learning without human annotations remains unexplored for mining materials and does not leverage on utilization of sensor-generated data. 
The proposed method, Depth Contrast, enables self-supervised learning of representations without labels on the 3DPM dataset by exploiting depth maps and inductive transfer. 
The proposed method outperforms material classification over ImageNet transfer learning performance in fully supervised learning settings and achieves an F1 score of 0.73. 
Further, The proposed method yields an F1 score of 0.65 with an 11\% improvement over ImageNet transfer learning performance in a semi-supervised setting when only 20\% of labels are used in fine-tuning. 
Finally, the Proposed method showcases improved performance generalization on linear evaluation. 
The implementation of proposed method is available on GitHub\footnote{https://github.com/prakashchhipa/Depth-Contrast-Self-Supervised-Method}.
\end{abstract}

\section{Introduction} \label{Introduction}


\begin{figure}[ht]
\centering
\includegraphics[width = 0.9\textwidth]{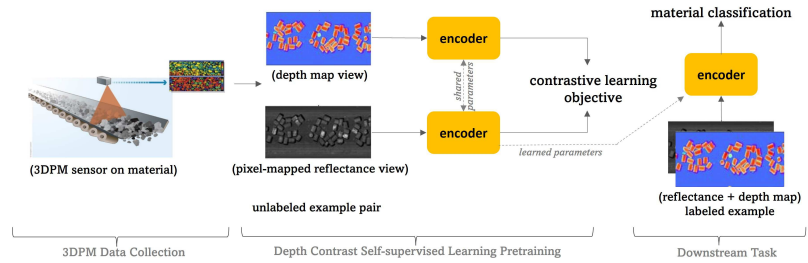}
\caption{The proposed approach comprises three steps: (1) Visual data collection of material on the conveyor belt through \ac{3DPM} sensor. It captures a depth map (height of material) and corresponding pixel-mapped reflectance data, the amount of light reflected from the imaged material registered by the camera sensor. (2) Self-supervised pre-training on unlabeled
pair of the depth map and reflectance visual data using proposed method \textbf{Depth Contrast} exploiting depth of material as supervision signal from data, thus enabling self-supervised representation learning beyond RGB images. (3) Performing the downstream task of material classification using the pretrained encoder where input is formulated by combining depth map and reflectance visual data. This work describes pixel-mapped reflectance data as reflectance image and depth maps as raw image in further sections.}
\label{fig:3dpm}
\end{figure}

Identifying and classifying materials on the conveyor belt is a fundamental process in the mining industry.
It is essential to properly sort the materials before forwarding it for stockpiling or further processing.
There is a need for real-time monitoring of the materials on the conveyor belt at mine sites for determining the destination for the material accurately and instantly. 
The traditional methods employed to identify materials or rock minerals are based on the physical and chemical properties of the material.
Also, based on human vision (basically human experience). 
In the recent past, advancement in the field of \ac{AI} and \ac{ML} has made it possible to successfully classify materials using data driven methods, more precisely intelligent computer vision methods \cite{cheng2017rock}, \cite{liu2019enhanced}, \cite{9042306}. This work aims to classify the materials using images from the \ac{3DPM} sensor \cite{3DPM} and transfer learning \cite{yosinski2014transferable}, chapter 15 of \cite{GoodBengCour16}, and employing \ac{DL} architectures. 
The deep \ac{CNNs} like Efficient-net \cite{tan2020efficientnet}, DenseNet \cite{huang2018densely}, ResNeXt \cite{xie2017aggregated} etc, are pretrained on large datasets such as ImageNet \cite{imagenet_cvpr09} and used to extract meaningful representations from new data. These high level representations are further used for classification. 

This work proposes a self-supervised representation learning (SSL) method Depth Contrast, to learn rich representations without human annotations to showcase the capabilities to significantly reduce the need for human supervision for downstream tasks, e.g., material classification. The Depth Contrast self-supervised representation learning method employs the temperature scaled cross-entropy loss for contrastive learning based on the SimCLR method \cite{chen2020simple}. The proposed pretraining method Depth Contrast exploits supervision signal from data, e.g., depth information of material, thus enabling to learn self-supervised representations on pixel-mapped reflectance data and depth maps from \ac{3DPM} sensor, which is beyond the typical RGB visual. 
Learned representations through SSL pretraining enable downstream tasks, e.g., material classifications with fewer human-annotated labeled examples, by leveraging domain-specific unlabeled data. Results of detailed experiments on downstream task material classification show that  1) It achieves benchmarked performance in mining material classification employing deep \ac{CNNs} with supervised ImageNet \cite{imagenet_cvpr09} transfer learning on the combined representation of reflectance and raw images 2) Proposed Depth Contrast method further improves the performance of material classification in a fully supervised setting and showcases the capabilities to reduced the needs of human annotation in a semi-supervised setting. The result section describes the detailed, quantified outcomes. 

\subsection{Measurement system} \label{3DPM}

3-Dimensional Particle Measurement (3DPM\textsuperscript{\tiny\textregistered}) \cite{3DPM} is a system that estimates the particle size distribution of material passing on a conveyor belt. 
3DPM can also determine particle shape, bulk volume flow and identify abnormalities such as the presence of boulders or rock bolts. 
Some benefits of online particle size distribution monitoring of bulk material on the conveyor are new control and optimization opportunities for grinding, crushing, and agglomerating. The system can be configured to individually estimate particle size distribution for a large variety of materials such as limestone, coke, agglomerated materials, metal ores, etc. In applications where different materials are transported on the same conveyor belt, the current material must be determined accurately for automated system configuration.
First part of the Fig. \ref{fig:3dpm} shows the schematic demonstration of the \ac{3DPM} system based on laser line triangulation, which produces 3D data with known real-world coordinates. The data used in this study comprises a depth map and corresponding pixel-mapped reflectance data. Reflectance data is the amount of light reflected from the imaged material registered by the camera sensor.
The \ac{3DPM} data can be visualized as images but are different from RGB/gray images; however, they can be treated in similar ways. Table~\ref{tab:sample_images} shows the reflectance (left) and depth-map (right) images of cylindrical shaped material in first row, respectively.

\subsection{Related work}

Many studies focused on similar material type or rock type identification using computer vision and machine learning techniques.
In the past 5-6 years, the \ac{DL} algorithms were used for material or rock classification using images, which yielded some remarkable results without the use of image pre-processing and feature extraction.
The article \cite{cheng2017rock} exploits deep \ac{CNNs} for identification of rock images from Ordos basin and achieves an accuracy of 98.5\%. 
In \cite{liu2019enhanced}, a comparison of \ac{SVM}, a histogram of oriented gradient based random forest model, and comprehensive \ac{DL} model (\ie Inception-V3 + Color model), for  mineral recognition in rocks indicates that the \ac{DL} model has the best performance.
A faster R-CNN model along with VGG16 for feature extraction is used in \cite{8964384} for identifying nine different rock types and also hybrid images of multiple rocks. 
The designed model gave an excellent accuracy of 96\% for independent rock types and around 80\% for hybrid rock type images.
Similar work in \cite{9042306} employs transfer learning methods using two light-weighted \ac{CNNs} SqueezeNet and MobileNets for classifying 28 types of rocks from images. 
These networks provide quick and accurate results on the test data; for example, SqueezeNet's accuracy is 94.55\% with an average recognition time of 557 milliseconds for a single image.
Since all these studies were carried out for different datasets and a different number of classes, it is not possible to compare them.
Also, a few of the above studies focus on images of ores, some on rocks, mineral rocks, and many other related variants.

Most of the previous works for ore type or material classification have used 2-dimensional imaging systems that work with RGB images or grayscale images.
The works which employ \ac{ML} or \ac{DL} are majorly carried out in supervised settings and do not exhibit much progress in utilizing the unlabeled data in either unsupervised or self-supervised settings. 
Self-supervised learning in computer vision attempts to learn representations from visuals without the need for annotations. 
There has been several approaches for self-supervised representation learning as (i) contrastive joint embedding methods (PIRL~\cite{misra2020self}  SimCLR~\cite{chen2020simple}, SimCLRv2~\cite{chen2020big}, and MoCo~\cite{he2020momentum}), (ii) quantization (SwAV~\cite{caron2020unsupervised} and DeepCluster~\cite{caron2018deep}), (iii) knowledge distillation (BYOL~\cite{grill2020bootstrap} and SimSiam~\cite{chen2021exploring}), and (iv) redundancy reduction (BT~\cite{zbontar2021barlow} and VICReg~\cite{bardes2021vicreg}).  
This work specifically focuses on self-supervised methods based on contrastive learning (CL).
The state-of-the-art contrastive learning methods SimCLR \cite{chen2020simple}, \cite{chen2020big} and MoCo~\cite{he2020momentum} learn by minimizing the distance between representations of different augmented views of the same visual example (`positive pairs') and maximizing the distance between the representations of augmented views from different visual examples(`negative pairs').
However, applying self-supervised learning to leverage on utilization of unlabeled data is not explored beyond the natural images. 
Therefore, the current work proposes contrastive learning based novel self-supervised pretraining method Depth Contrast for learning representations on visual data (raw and reflectance images) from \ac{3DPM} sensor and further improving the performance of downstream task, online material type classification model from the \ac{3DPM} images while material passes on the conveyor belt. 
Many \ac{DL} based models are investigated, and transfer learning is used to improve the task performance.

The rest of the paper is organized as follows. 
Section \ref{data} discusses the dataset used in this study. 
Methodology and experiments are discussed in Section \ref{Methodology and experiments}. 
Section \ref{Result and Discussion} presents the results of the proposed work. 
Finally, we conclude in Section \ref{Conclusion and Future Scope}.

\section{3DPM Dataset} \label{data}
For material classification, there are two modalities of data that are acquired using the \ac{3DPM} sensor. 
First are the reflectance images, which are 2D grayscale images of the material on the conveyor belt taken in red laser light (680 nm). 
Lighting conditions, color, and distance between the sensor and the conveyor belt \ie base depth may affect the 2D imaging system \cite{thurley2009automated}.  
The second modality is the 3D depth map of the bulk material, \ie the distance between the material and the sensor.
As the material is passed on the conveyor belt, data of multiple laser lines are collected at regular intervals, thereby creating a 3D depth image. 
2D imaging systems have scale or perception bias, which is resolved in this 3D data as it is unaffected by the change in light and color of the material \cite{THURLEY2011254}. 
The 3DPM dataset is prepared for this work by collecting the data from sensors under similar lighting conditions. Further, the challenge of variable base depth, which is an issue for the depth map, is solved by subtracting the conveyor depth from all the data. 
Therefore, every reflectance image has a corresponding raw image (\ie depth maps). 
Table~\ref{tab:sample_images} shows the sample \ac{3DPM} images of all the categories of material in the 3DPM dataset.
In order to feed these images to the \ac{CNNs} three arrangements were made, only raw images, only reflectance images, and respective reflectance and raw images are merged, forming a three-channel image. 
The experiments proved that combining the reflectance and raw images led to higher performance, refer to Table \ref{tab:exp}.

The 3DPM dataset has seven classes corresponding to the categories of mining material available on the conveyor belt. These classes are Cylindrical, Ore1, Ore2, Ore3, Mixed1, Mixed2, and Agglomerated (Aggl.). Some of the material categories belong to a specific type of material e.g., Cylindrical, Ore1, Ore2, Ore3, and Agglomerated, whereas the remaining categories, such as Mixed1 and Mixed2, represent the mixture of several materials present on the conveyor belt. Identification of Mixed categories remains challenging with manual visual inspections. The class distribution is shown in Table.~\ref{tab:data_dist}.

\begin{table}[]
\centering
\caption{Classwise raw and reflectance image pairs}
\label{tab:sample_images}

\resizebox{0.8\textwidth}{!}{%
\begin{normalsize}
\begin{tabular}{cccccc}
Material    & Reflectance Image & Raw Image & Material & Reflectance Image & Raw Image \\
\rotatebox[origin=l]{90}{\hspace{1.5em}Cylindrical} & \includegraphics[width=.45\linewidth]{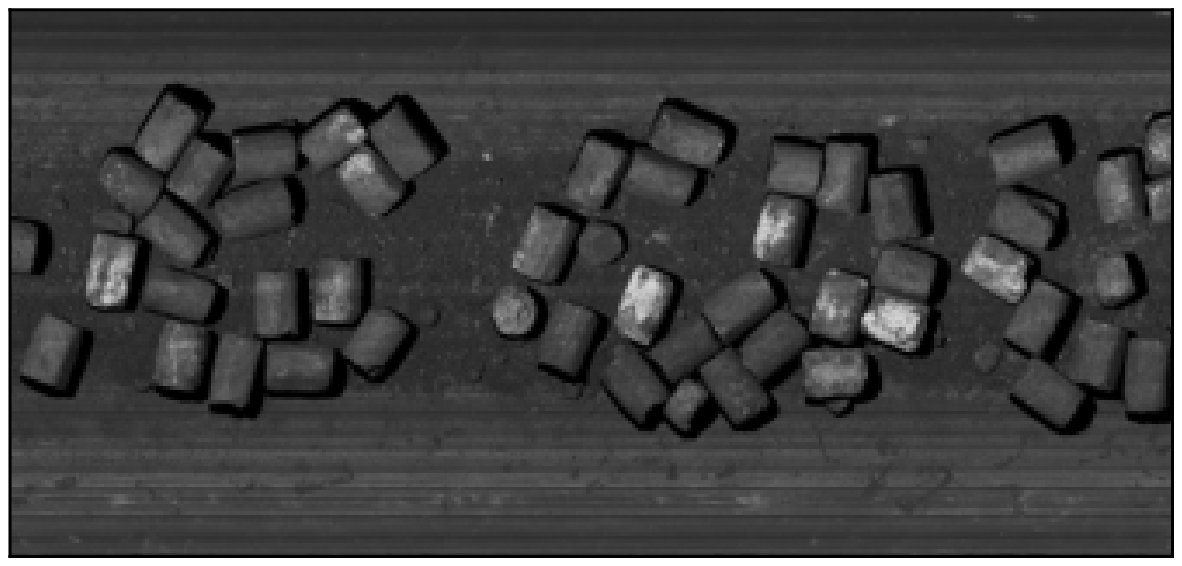} & \includegraphics[width=.45\linewidth]{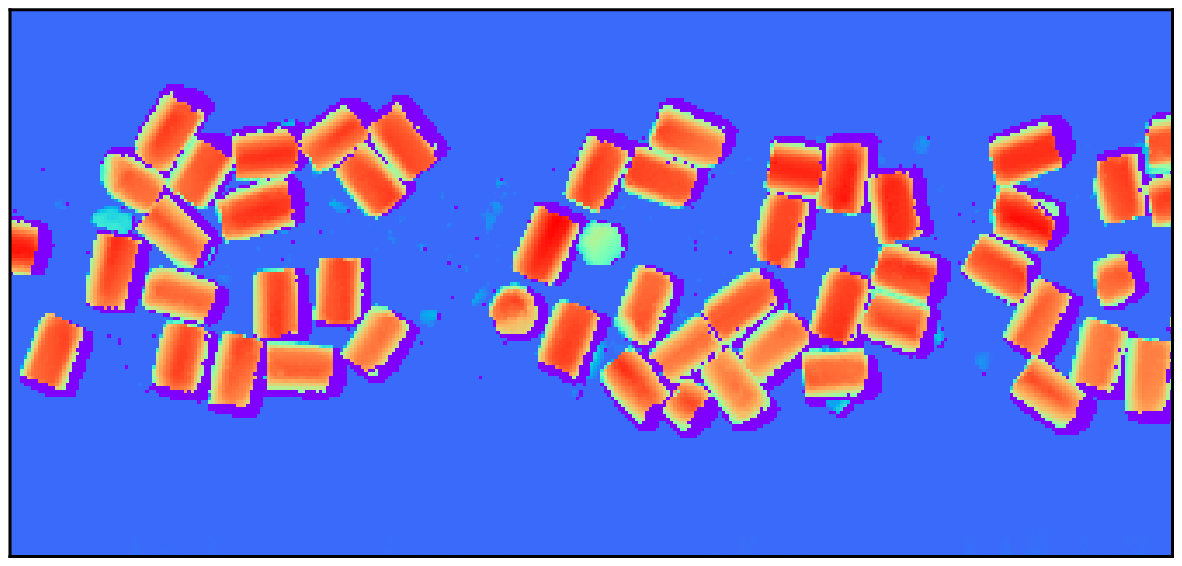}           & \rotatebox[origin=l]{90}{\hspace{1.5em} Mixed1} & \includegraphics[width=.45\linewidth]{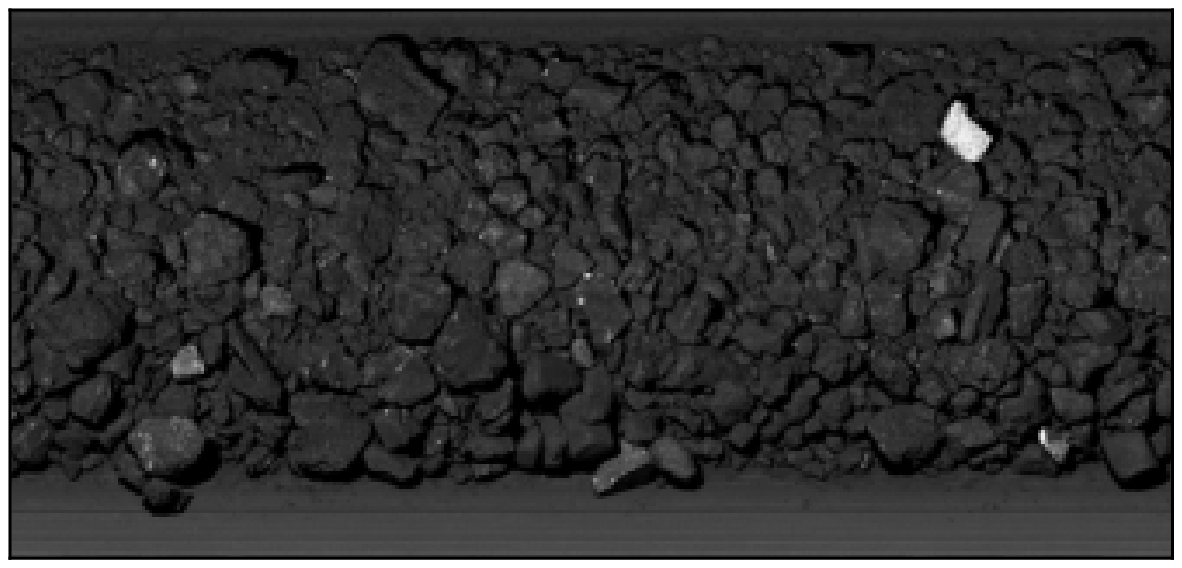} &
     \includegraphics[width=.45\linewidth]{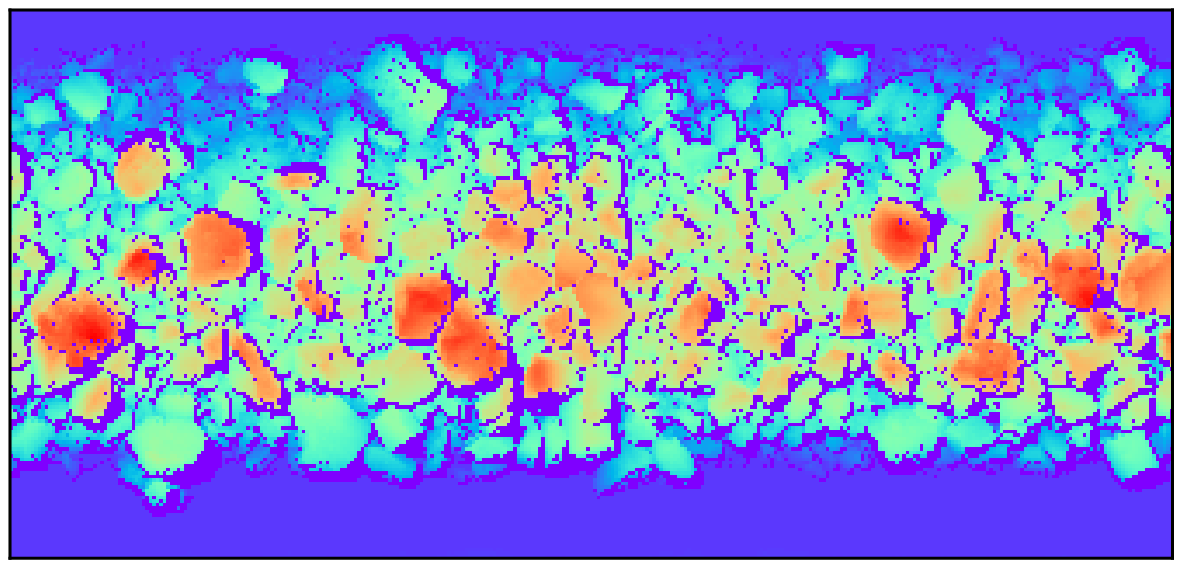}         \\
\rotatebox[origin=l]{90}{\hspace{1.5em} Ore1} & \includegraphics[width=.45\linewidth]{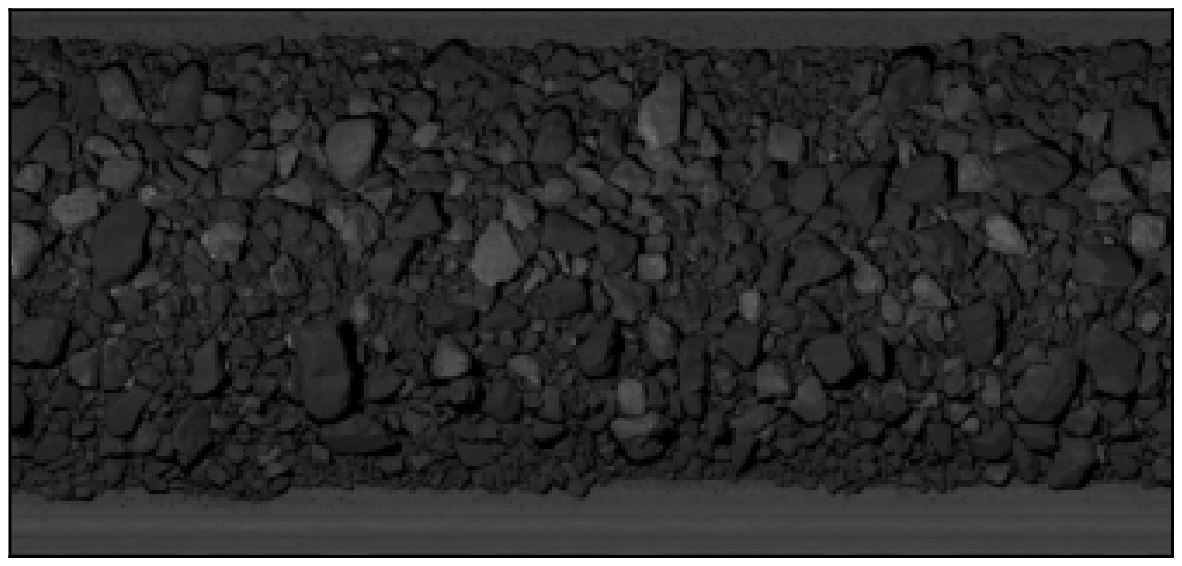} & \includegraphics[width=.45\linewidth]{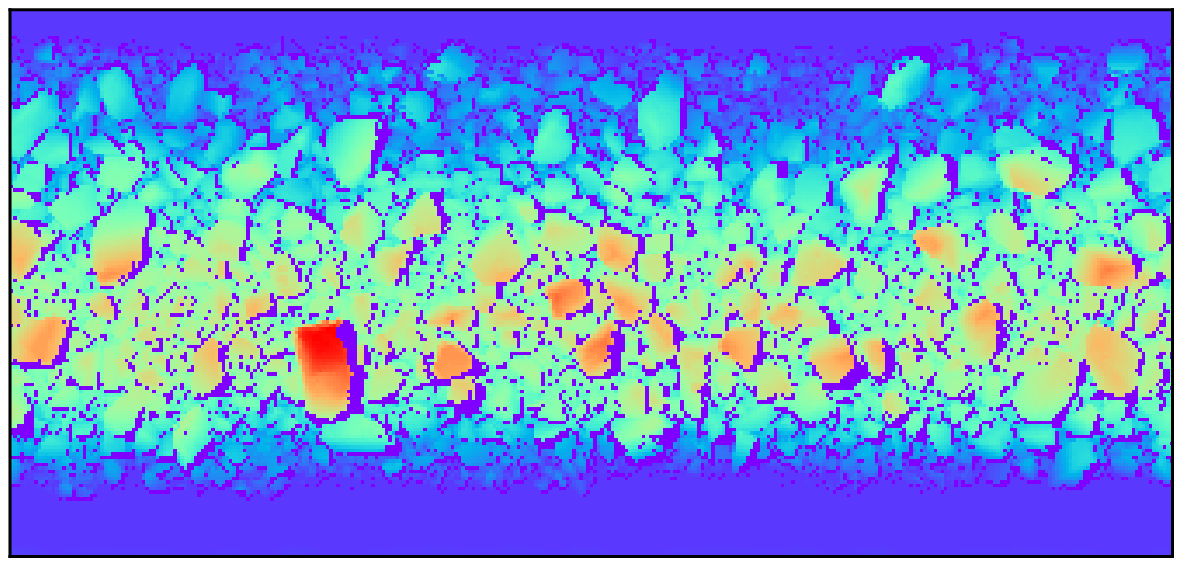}           & \rotatebox[origin=l]{90}{\hspace{2em} Aggl.} & \includegraphics[width=.45\linewidth]{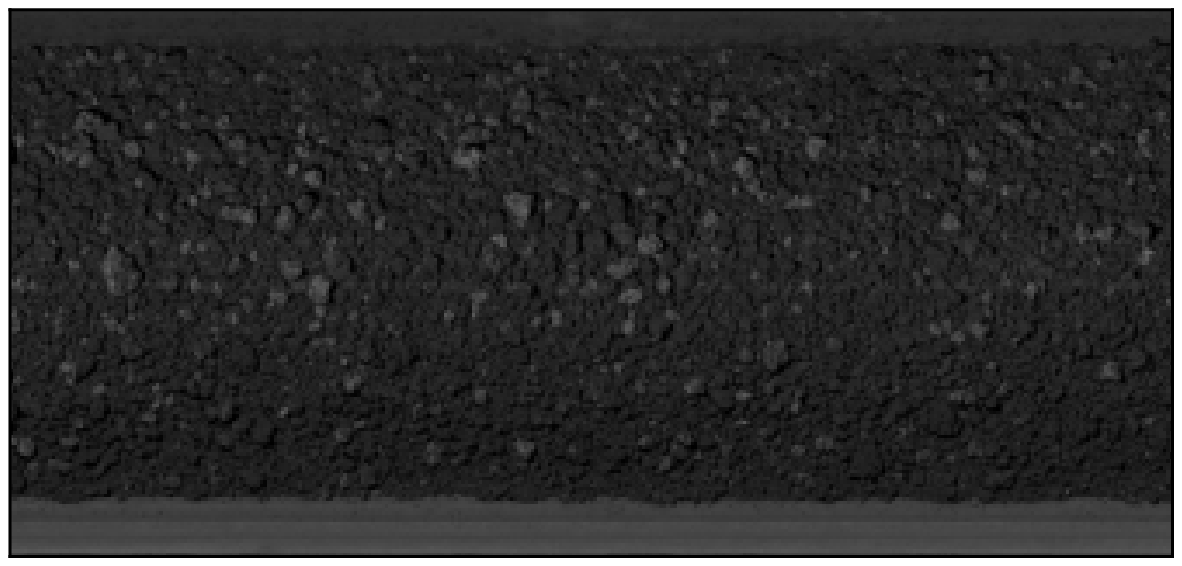} & \includegraphics[width=.45\linewidth]{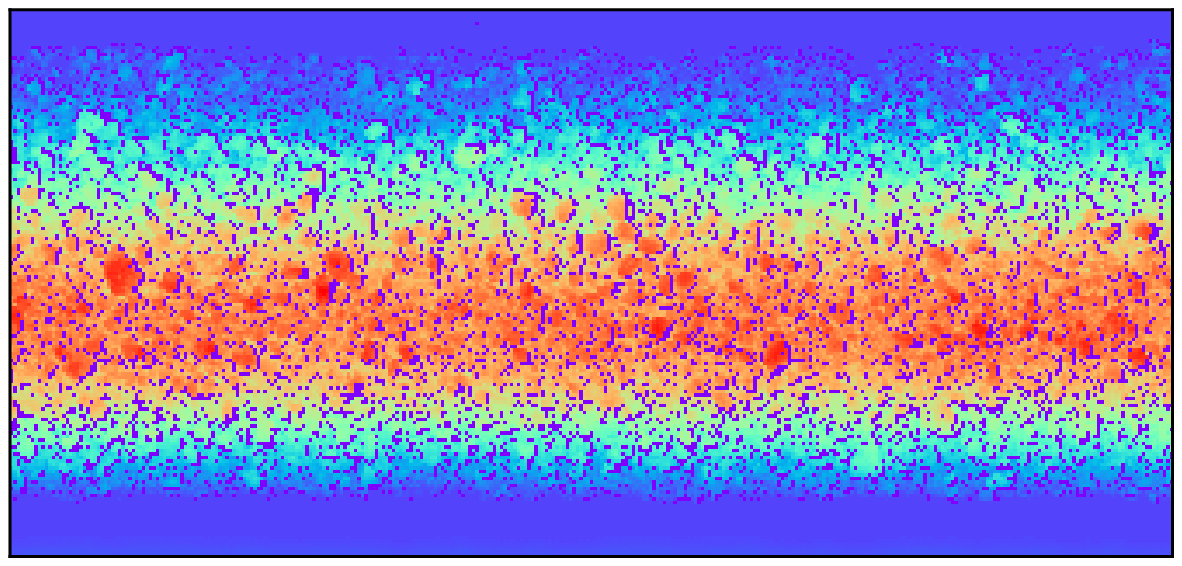}           \\
\rotatebox[origin=l]{90}{\hspace{1.5em}Ore2} & \includegraphics[width=.45\linewidth]{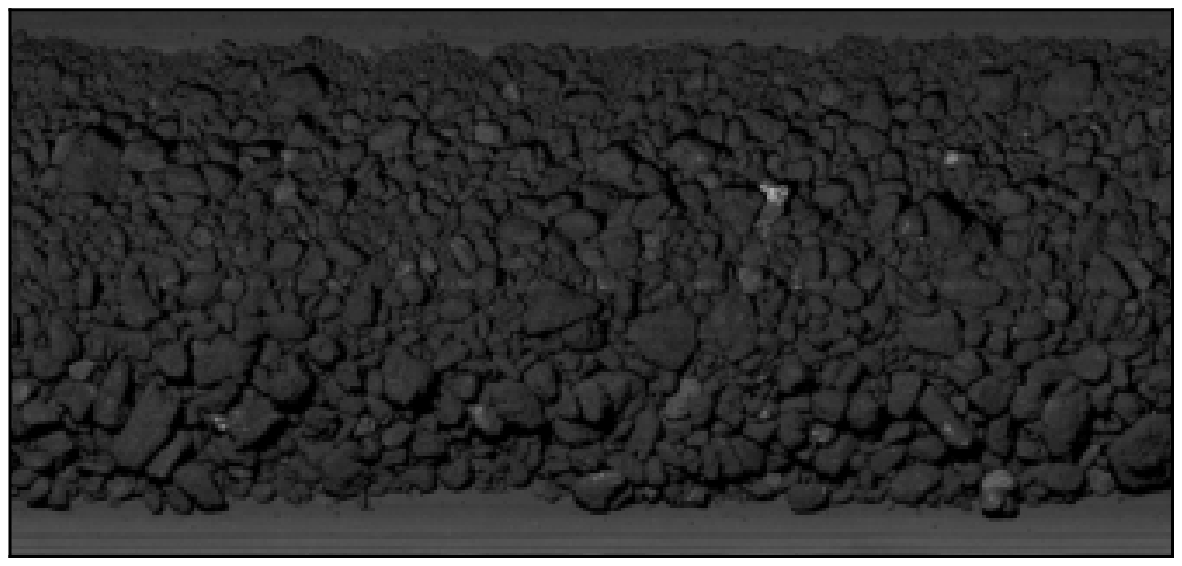} & \includegraphics[width=.45\linewidth]{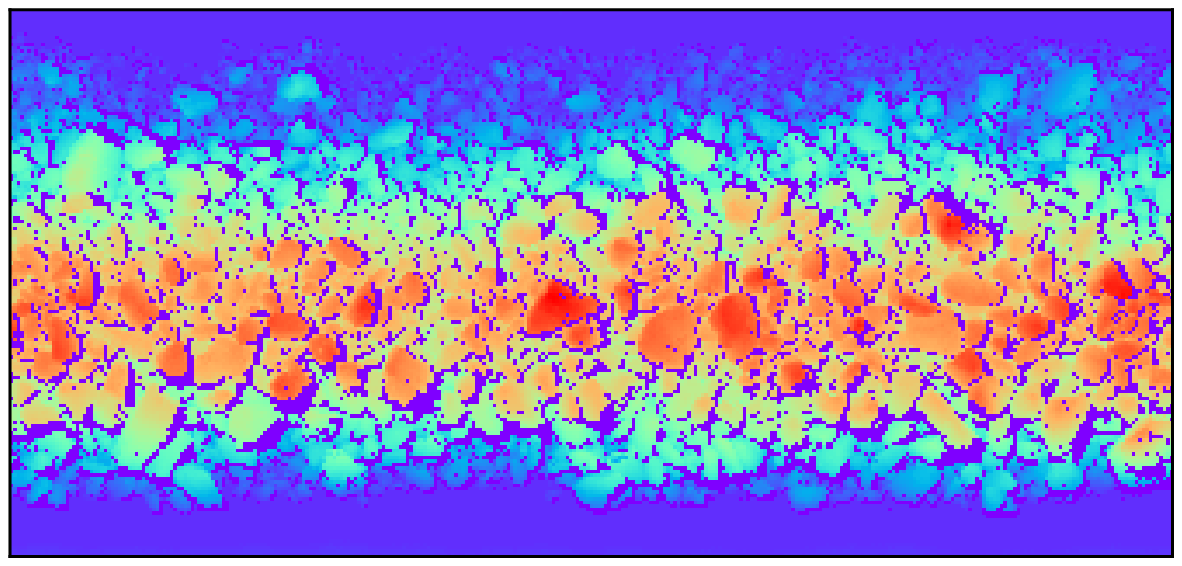}           & \rotatebox[origin=l]{90}{\hspace{2em} Mixed2} & \includegraphics[width=.45\linewidth]{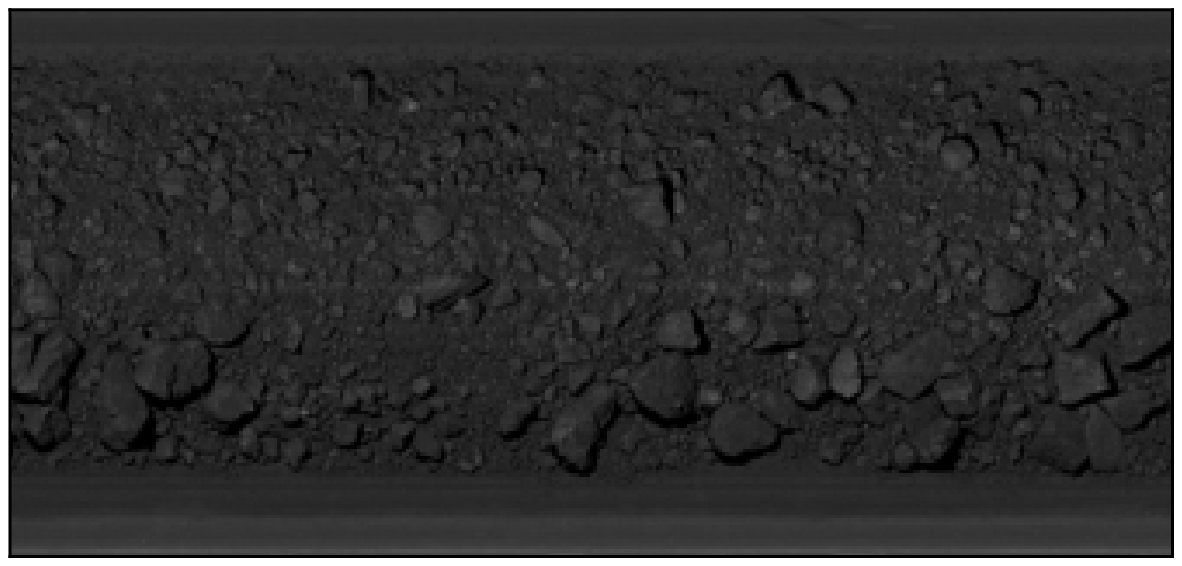} & \includegraphics[width=.45\linewidth]{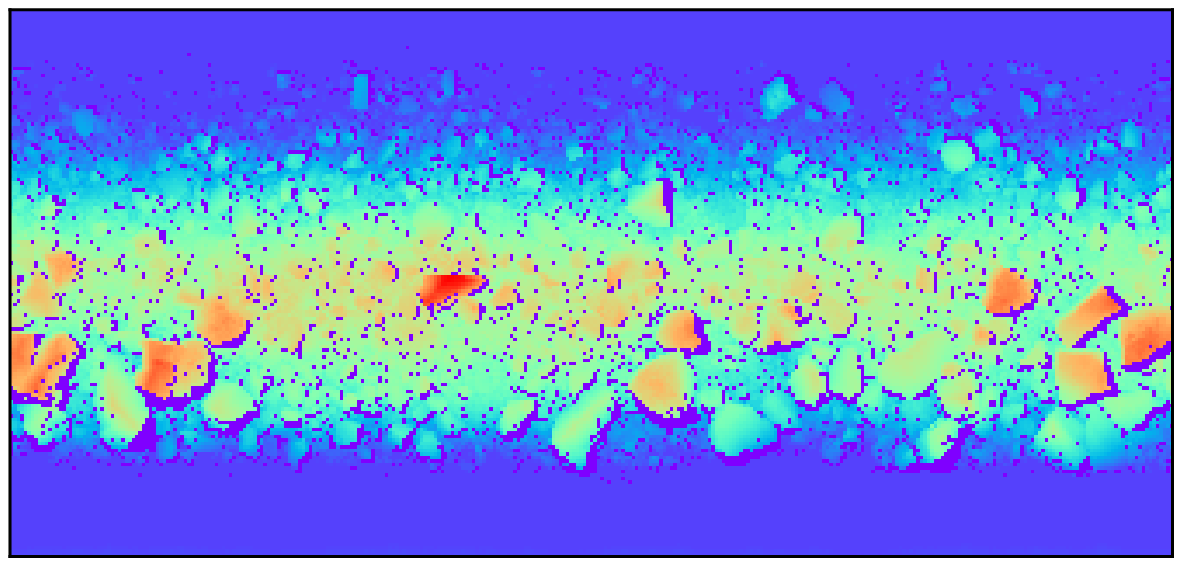}           \\
\rotatebox[origin=l]{90}{\hspace{2.5em} Ore3}  & \includegraphics[width=.45\linewidth]{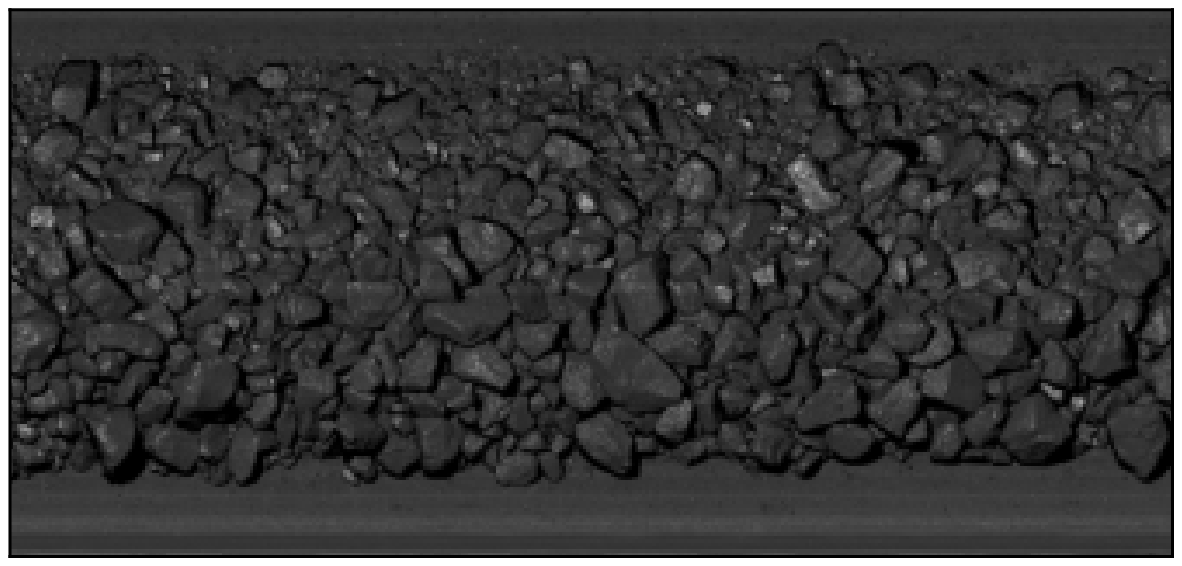} & \includegraphics[width=.45\linewidth]{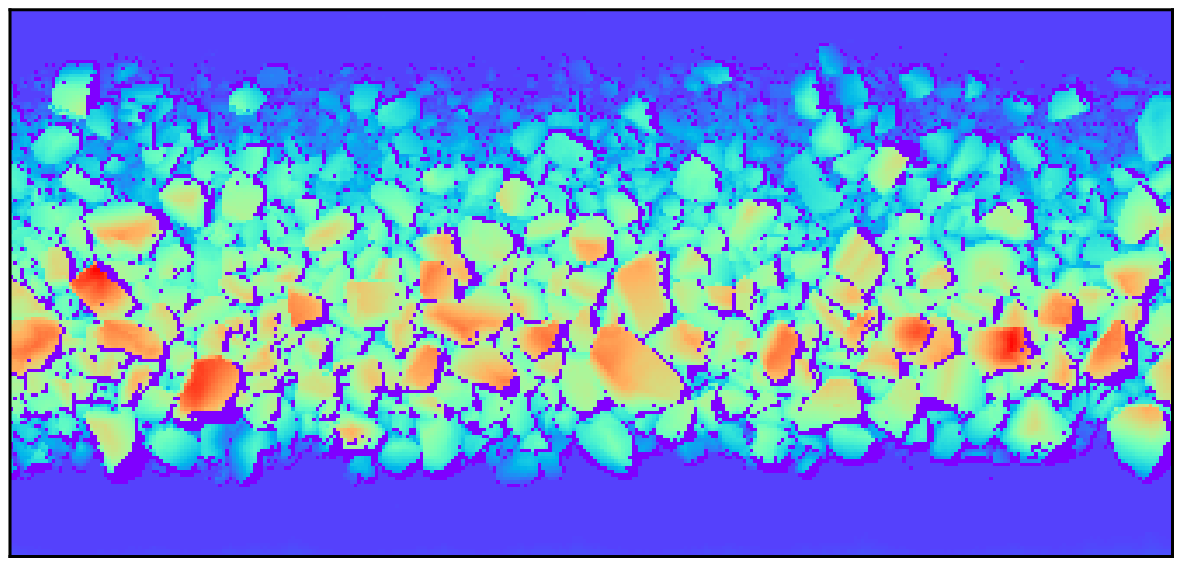}           &          &                   &          
\end{tabular}
\end{normalsize}
}
\end{table}


\begin{table}[!ht]
\centering
\caption{Types of materials and their distribution}
\label{tab:data_dist}
\scriptsize
\begin{tabular}{lc}
\hline
\textbf{Material name} & \textbf{Total images} \\ \hline

Mixed 1 & 164 \\
Mixed 2 & 122 \\
Ore 1 & 860 \\
Ore 2 & 698 \\
Ore 3 & 503 \\
Agglomerated (Aggl.) & 616 \\
Cylindrical & 45 \\ \hline
\end{tabular}%

\end{table}

\section{Methodology and Experiments}\label{Methodology and experiments}

This work aims to investigate the classification of materials using supervised and self-supervised learning methods. 
For faster and more efficient training with the limited (insufficient) data which is available at hand, transfer learning \cite{9042306} is employed to derive essential features to achieve better classification performance.
This section discusses transfer learning and the supervised and self-supervised techniques used in this work.

\subsection{Transfer Learning} \label{transferlearning}
Transfer learning \cite{tan2018survey} is a method in \ac{ML} that exploits the existing knowledge of a network to learn a distinct but relevant task efficiently. 
In transfer learning, the network is first well-trained on a very large dataset to solve a specific problem (maybe classification, regression, etc.). This mechanism is referred to as the source task.
For another task \ie target task, which has lesser data than the source task, the knowledge acquired by the source task is shared in network parameters (or weights) to accomplish the target task. 
The transfer of information can be carried out in two ways; the first is when the learned network parameters of the source task, act as a good starting point for the target task, and the complete network can further be fine-tuned while training on the target data, refereed as `fine-tuning' in this article. 
The second is when the source network is used as it is to extract features for the target task, referred to as `linear evaluation'.
In this work, transfer learning is applied in both supervised and self-supervised settings.
Already trained networks on ImageNet \cite{imagenet_cvpr09}, which is a substantial multi-label image database, are utilized here for material type classification.
These are further discussed in section \ref{supervised} and section \ref{selfsupervised}.

\subsection{Supervised Downstream Task} \label{supervised}
Supervised learning \cite{GoodBengCour16} is a type of \ac{ML} technique that uses human-annotated labels to predict the outcome of the learning problem. As the 3DPM dataset has labels along with the reflectance and raw images containing depth maps of the materials, thus applying supervised learning methods for downstream tasks of classification is reasonable to be applied in both supervised transfer learning section ~\ref{transferlearning} as well as self-supervised Depth Contrast learning section ~\ref{selfsupervised}. 
This work considers three pretrained networks \ie Efficient-net b2\cite{tan2020efficientnet}, DenseNet-121\cite{huang2018densely} and ResNeXt-50 (32$\times$4d)\cite{xie2017aggregated} for ImageNet\cite{imagenet_cvpr09} transfer learning method and EfficientNet-b2\cite{tan2020efficientnet} for self-supervised method for representation learning using Depth Constrast method. 
Since the downstream task of material classification on the 3DPM dataset has seven classes corresponding to 7 categories of materials thus, the above-mentioned CNN networks are extended with a fully connected (FC) layer module having two FC layers, one hidden layer of 512, and one output layer of 7 units.   



\subsection{Depth Contrast - Self-supervised learning} \label{selfsupervised}
\begin{figure}[h!t]
    \centering
    \includegraphics[scale = 0.45]{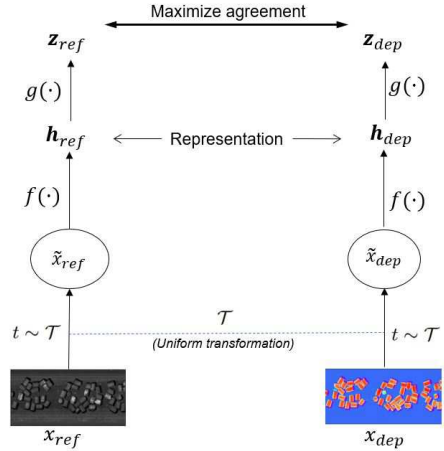}
    \caption{Depth Contrast method. An uniform data augmentation operator is sampled from one random-crop transform \textit{(t $\sim$ T)}, applied to each pair of both views, reflectance and depth image from \ac{3DPM} sensor, considering correlated views. A base encoder network \textit{f}(·) which is Efficient-net b2 and a projection head which is perceptron network \textit{g}(·) are trained to maximize agreement using a normalized temperature-scaled cross entropy loss from SimCLR \cite{chen2020simple}. Upon completion of pre-training, encoder network f(·) and representation \textbf{h} used for downstream classification task}
    \label{fig:depth_contrast_fig}
\end{figure}

With a moderate amount of data availability in terms of reflectance and depth images as pair, we propose a method, ``depth contrast", based on normalized temperature-scaled cross-entropy loss from SimCLR \cite{chen2020simple}. It models the two transformed views of input images from a pair of reflectance and depth images, respectively, and intends to maximize the agreement between the stated views of the same data example through contrastive loss. Derived from SimCLR \cite{chen2020simple}, following are the components for Depth Contrast, mentioned in Figure \ref{fig:depth_contrast_fig}.
\begin{itemize}
    \item A \textit{domain specific prior} and \textit{stochastic transformation} based module that transforms the synchronously yet independently generated pair of reflectance and depth views of the same example through uniform transformation of random cropping operation, denoted as \textbf{$\tilde{x}_{ref}$} and \textbf{$\tilde{x}_{dep}$} which are considers as positive pair. Both, reflectance and depth views are 32 bit images and depth view specifically captures the 3D depth of the material.
    \item  A neural network \textit{base encoder} \textit{f}(·) which yields representations from transformed reflectance and depth views of data examples. Specifically, this work uses Efficient-net b2 \cite{tan2020efficientnet} as base encoder based on empirical analysis of CNNs in transfer learning based supervised learning method and further obtains obtain $h_{ref}$ = \textit{f}($\tilde{x}_{ref}$) = Efficient-net($\tilde{x}_{ref}$) and $h_{dep}$ = \textit{f}($\tilde{x}_{dep}$) = Efficient-net($\tilde{x}_{dep}$) where $h_{ref}$, $h_{dep}$ $\in$ \& $\mathbb{R}^d$ are the output after the average pooling layer respectively.
    \item A small-scale multi-layer perceptron (MLP) network projection head \textit{g}(·) that maps representations to the latent space where contrastive loss is applied. A MLP having three hidden layers of 2048, 2048, and 512 neurons and output layer with 128 neurons to obtain $z_{ref}$ = \textit{g}($h_{ref}$) = $W^{(2)}$$\sigma$$(W^{(1)}{h_{ref})}$ and $z_{dep}$ = \textit{g}($h_{dep}$) = $W^{(2)}$$\sigma$$(W^{(1)}{h_{dep})}$ where $\sigma$ is a ReLU non-linearity.
    \item A contrastive loss function, normalized temperature-scaled cross entropy loss (NT-Xent) from SimCLR \cite{chen2020simple} is defined for a contrastive prediction task. For given a set \textbf{$\tilde{x}_{k}$} including a positive pair of examples \textbf{$\tilde{x}_{ref}$} and \textbf{$\tilde{x}_{dep}$} , the contrastive prediction task intend to identify \textbf{$\tilde{x}_{dep}$} in \{${\textbf{$\tilde{x}_{k}$}}\}_{k\neq ref}$ for a given \textbf{$\tilde{x}_{ref}$}.
\end{itemize}
The loss function for a positive pair of examples (ref, dep) is defined as
\begin{equation} \label{eq:1}
    L_{ref,dep} = -log \dfrac{exp(sim(\boldsymbol{z}_{ref},\boldsymbol{z}_{dep})/\tau)} {\sum_{k =1}^{2N} 1_{[k \neq ref]} exp(sim(\boldsymbol{z}_{ref},\boldsymbol{z}_{k})/\tau)}
\end{equation}



In equation \ref{eq:1}, where $1_{[k \neq ref]} \in {0, 1}$ is an indicator evaluating to 1 if $k \neq i$. Equations 2 and 3 define dot product between $l_2$ normalized respective quantities, and $\tau$ defines the temperature parameter. During pre-training, a randomly sampled minibatch of N examples is randomly sampled. The contrastive prediction task is defined on pairs of reflectance and depth views derived from the minibatch, resulting in 2N data points. A sampling of negative examples is not performed explicitly. Instead, given a positive pair, the other 2(N $-$ 1) augmented examples within a minibatch are considered negative examples. The loss value is computed across all positive pairs of reflectance and depth views, both $(\boldsymbol{z}_{ref},\boldsymbol{z}_{dep})$ and $(\boldsymbol{z}_{dep},\boldsymbol{z}_{ref})$, in each mini-batch.

\subsection{Experiments}

The experiments are designed for in-depth analysis of both types of input images, \ac{CNNs} architectures, and comparative analysis of proposed self-supervised representation learning method Depth Contrast with ImageNet\cite{imagenet_cvpr09} pretrained models on a different level of labeled data availability.
Broadly, training of material classification tasks is defined in two arrangements, fine-tuning and linear evaluation,
respectively. 


\begin{table*}[ht]
\centering
\caption{Performance comparison between ImageNet pretrained and Depth Contrast pretrained models \textbf{fine-tuned} on \ac{3DPM} dataset in semi and fully supervised setting }
\scriptsize
\begin{tabular}{cccccclc}
\hline
Exp. No. & Input & Model & \multicolumn{3}{c}{Dataset split (in \%)} & \multicolumn{2}{l}{Performance  (in F1 score)} \\ \cline{4-8} 
 & \multicolumn{1}{l}{} & \multicolumn{1}{l}{} & \multicolumn{1}{l}{Train} & \multicolumn{1}{l}{Val.} & \multicolumn{1}{l}{Test} & \multicolumn{1}{c}{Train} & Test \\ \hline
 & \multicolumn{7}{c}{\textbf{ImageNet pre-trained}} \\
1 & Raw images & {\color[HTML]{000000} EfficientNet} & {\color[HTML]{000000} 60} & {\color[HTML]{000000} 20} & {\color[HTML]{000000} 20} & 0.31 ± 0.0142 & 0.30 ± 0.0061 \\
2 & Reflectance images & {\color[HTML]{000000} EfficientNet} & {\color[HTML]{000000} 60} & {\color[HTML]{000000} 20} & {\color[HTML]{000000} 20} & 0.68 ± 0.0081 & 0.69 ± 0.0235 \\
 &  & {\color[HTML]{000000} 3.1 EfficientNet} & {\color[HTML]{000000} 60} & {\color[HTML]{000000} 20} & {\color[HTML]{000000} 20} & 0.73 ± 0.0077 & \textbf{0.72 ± 0.0294} \\
3 & Raw + Reflectance & {\color[HTML]{000000} 3.2 ResNeXt} & {\color[HTML]{000000} 60} & {\color[HTML]{000000} 20} & {\color[HTML]{000000} 20} & 0.75 ± 0.0321 & 0.67 ± 0.0841 \\
 &  & {\color[HTML]{000000} 3.3 DenseNet} & {\color[HTML]{000000} 60} & {\color[HTML]{000000} 20} & {\color[HTML]{000000} 20} & 0.74 ± 0.0297 & 0.72 ± 0.0411 \\
4 & Raw + Reflectance & EfficientNet & 10 & 10 & 20 & 0.65 ± 0.0616 & 0.54 ± 0.0231 \\
 & \multicolumn{7}{c}{\textbf{Depth Contrast pre-trained (60\% unlabelled data)}} \\
5 & Raw + Reflectance & {\color[HTML]{000000} EfficientNet} & {\color[HTML]{000000} 60} & {\color[HTML]{000000} 20} & {\color[HTML]{000000} 20} & 0.73 ± 0.0152 & \textbf{0.73 ± 0.0135} \\
6 & Raw + Reflectance & {\color[HTML]{000000} EfficientNet} & {\color[HTML]{000000} 10} & {\color[HTML]{000000} 10} & {\color[HTML]{000000} 20} & 0.71 ± 0.0399 & \textbf{0.65 ± 0.0263} \\ \hline
\end{tabular}
\label{tab:exp}
\end{table*}
In fine-tuning training task, all the layers of \ac{CNN} architecture and fully connected layers are trained on \ac{3DPM} dataset (refer to Table \ref{tab:exp} whereas in the linear evaluation training task mentioned in Table \ref{tab:exp_le}, only fully connected layers are trainable and  \ac{CNN} architecture serve as a feature extractor. 


\begin{table*}[ht]
\centering
\caption{Performance comparison between ImageNet pretrained and Depth Contrast pretrained models \textbf{linear evaluation} on \ac{3DPM} dataset in semi and fully supervised setting}
\scriptsize
\begin{tabular}{cccccclc}
\hline
Exp. No. & Input & Model & \multicolumn{3}{c}{Dataset split (in \%)} & \multicolumn{2}{l}{Performance  (in F1 score)} \\ \cline{4-8} 
 & \multicolumn{1}{l}{} & \multicolumn{1}{l}{} & \multicolumn{1}{l}{Train} & \multicolumn{1}{l}{Val.} & \multicolumn{1}{l}{Test} & \multicolumn{1}{c}{Train} & Test \\ \hline
 & \multicolumn{7}{c}{\textbf{ImageNet pre-trained}} \\
7 & Raw + Reflectance & EfficientNet & 60 & 20 & 20 & 0.74 ± 0.0825 & 0.69 ± 0.0431 \\
8 & Raw + Reflectance & EfficientNet & 10 & 10 & 20 & 0.67 ± 0.0364 & 0.52 ± 0.0380 \\
 & \multicolumn{7}{c}{\textbf{Depth Contrast pre-trained (60\% unlabelled data)}} \\
9 & Raw + Reflectance & {\color[HTML]{000000} EfficientNet} & {\color[HTML]{000000} 60} & {\color[HTML]{000000} 20} & {\color[HTML]{000000} 20} & 0.74 ± 0.0273 & \textbf{0.72 ± 0.0854} \\
10 & Raw + Reflectance & {\color[HTML]{000000} EfficientNet} & {\color[HTML]{000000} 10} & {\color[HTML]{000000} 10} & {\color[HTML]{000000} 20} & 0.71 ± 0.0624 & \textbf{0.64 ± 0.0147} \\ \hline
\end{tabular}
\label{tab:exp_le}
\end{table*}
Comparative study of these two types of tasks allows for analyzing representation learning capabilities of the proposed method Depth Contrast in a more detailed manner.
This work uses a 5-cross stratified validation technique in which each fold comprises 20\% data ensuring class distributions to ensure robustness in performance.
In general 60\% data, equal to three folds used as the train set, 20\%, one fold for the validation set, and the remaining 20\%, one fold for the test set. This procedure is repeated five times to test each fold at once, and variations in results are reported in terms of standard deviation alongside the mean F1 score.   

Fine-tuning experiments mentioned in Table \ref{tab:exp}, Experiment 1, 2, and 3.1 are focused on investigating the effectiveness of input image type for material classification and chooses Efficient-net b2 as base architecture based on preliminary analysis on several CNN architectures. 
Further, experiments 3.2 and 3.3 focus on exploring different CNN architectures in which input types remain a combination of raw and reflectance images, based on performance analysis of previous experiments. Based on preliminary empirical analysis, the combination of raw and reflectance images is composed of three-channel images and raw-reflectance-raw channels. Specifically, experiment 3.2 uses DenseNet 121 \cite{huang2018densely} and experiment 3.3 uses ResNeXt (31 $\times$ 4d) \cite{xie2017aggregated} architecture. 

The purpose of experiment 4 in Table \ref{tab:exp} is to benchmark the material classification performance in a limited labeled data setting while using the Efficient-net network over a combination of raw and reflectance images based on the performance of previous experiments. This experiment uses a single fold where 10\% data for training and the remaining 10\% data for validation. However, the test set remains the same for all folds. It is worth noticing that experiments 1 to 4 uses ImageNet\cite{imagenet_cvpr09} pretrained models.

Finally, Experiments 5 and 6 in Table \ref{tab:exp} are focused on evaluating the effect of the proposed self-supervised representation learning method, Depth Contrast, in semi and fully supervised settings, respectively. 
Before these experiments, the Efficient-net b2 model is pretrained using the Depth Contrast method over training data without using labels. 
This pretraining allows the model to learn domain-specific representations in a self-supervised manner. 
Experiment 5 uses training data the same as previous experiments 1 to 3. 
Another side, experiment 6 evaluates the performance in a limited labeled data set to compare the effectiveness of Depth Contrast pretraining with experiment 4, while both the experiments use weights initialization from the Depth Contrast method

Experiments dedicated to linear evaluation training tasks are mentioned in Table \ref{tab:exp_le}. Experiment 7 and 8 investigates the performance on ImageNet \cite{imagenet_cvpr09} pretrained model corresponds to fine-tuning experiments 3.1 and 4. Similarly, experiments 9 and 10 correspond to fine-tuning experiments 5 and 6 to investigate the performance of the Depth Contrast pretrained model.  
In a nutshell, a comparative analysis of the aforementioned experiments can analyze the effectiveness of different transfer learning mechanisms and the leverage of unlabelled sensor-generated data to improve performance in typical industrial settings. 

All the fine-tuning experiments mentioned in Table \ref{tab:exp} and linear evaluation experiments in Table \ref{tab:exp_le} for material classification follow the \textit{learning rate} of 0.00001, \textit{Adam optimizer}, \textit{batch size} of 16, \textit{input image size} of $224\times224$, and \textit{dropout regularization} of $0.3$ for the hidden layer of FC layer module. Images are 32 bit, so current work only uses random crop as an augmentation technique. Further, Depth Contrast pretraining method uses the same input image size with uniform random crop to both input views, \textit{batch size} of 256, a \textit{learning rate} of 0.00005, \textit{temperature parameter} value of 0.1, and the \textit{MLP projector head} comprises three hidden layers with 2048-2048-512 neurons with ReLU and a batch normalization layer. The output layer of the projector head consists of 128 neurons during Depth Contrast pre-training. The architecture of the MLP projection head is chosen based on preliminary empirical analysis.
Further, the F-score evaluation metric is used to accurately measure the performance of models while having class imbalance presents. The empirical analysis of network architectures and training parameters are confined by limited human \& computing resources. Methodological development for Depth Contrast and its investigation remains the main focus.

\section{Result and Discussion}\label{Result and Discussion}
\begin{table*}[ht]
\centering
\caption{Class-wise train and test performance for ImageNet pretrained and fine-tuned on \ac{3DPM} dataset in fully supervised setting}
\tiny
\begin{tabular}{cccccccc}
\hline
 &  & \multicolumn{6}{c}{Experiment No. 3.1} \\ \cline{3-8} 
S.No. & Classes & \multicolumn{3}{c}{Train} & \multicolumn{3}{c}{Test} \\ \cline{3-8} 
 &  & Precision & Recall & F1 & Precision & Recall & F1 \\ \hline
1 & Mixed 1 & {\color[HTML]{1E1E1E} 0.31 ± 0.0235} & {\color[HTML]{1E1E1E} 0.63 ± 0.1522} & {\color[HTML]{1E1E1E} 0.41 ± 0.0457} & {\color[HTML]{1E1E1E} 0.31 ± 0.0589} & {\color[HTML]{1E1E1E} 0.61 ± 0.1846} & {\color[HTML]{1E1E1E} 0.40 ± 0.0643} \\
2 & Agglomerated & {\color[HTML]{1E1E1E} 0.65 ± 0.0755} & {\color[HTML]{1E1E1E} 0.84 ± 0.0166} & {\color[HTML]{1E1E1E} 0.73 ± 0.0539} & {\color[HTML]{1E1E1E} 0.64 ± 0.0499} & {\color[HTML]{1E1E1E} 0.84 ± 0.0425} & {\color[HTML]{1E1E1E} 0.72 ± 0.0203} \\
3 & Mixed 2 & {\color[HTML]{1E1E1E} 0.70 ± 0.2032} & {\color[HTML]{1E1E1E} 0.23 ± 0.0405} & {\color[HTML]{1E1E1E} 0.34 ± 0.0488} & {\color[HTML]{1E1E1E} 0.63 ± 0.3924} & {\color[HTML]{1E1E1E} 0.22 ± 0.1996} & {\color[HTML]{1E1E1E} 0.29 ± 0.2164} \\
4 & Ore 1 & {\color[HTML]{1E1E1E} 0.89 ± 0.0185} & {\color[HTML]{1E1E1E} 0.72 ± 0.0527} & {\color[HTML]{1E1E1E} 0.79 ± 0.0320} & {\color[HTML]{1E1E1E} 0.89 ± 0.0113} & {\color[HTML]{1E1E1E} 0.72 ± 0.0537} & {\color[HTML]{1E1E1E} 0.80 ± 0.0294} \\
5 & Ore 2 & {\color[HTML]{1E1E1E} 0.89 ± 0.0721} & {\color[HTML]{1E1E1E} 0.72 ± 0.0433} & {\color[HTML]{1E1E1E} 0.79 ± 0.0371} & {\color[HTML]{1E1E1E} 0.89 ± 0.0841} & {\color[HTML]{1E1E1E} 0.71 ± 0.0284} & {\color[HTML]{1E1E1E} 0.79 ± 0.0300} \\
6 & Ore 3 & {\color[HTML]{1E1E1E} 0.69 ± 0.0768} & {\color[HTML]{1E1E1E} 0.71 ± 0.0450} & {\color[HTML]{1E1E1E} 0.70 ± 0.0528} & {\color[HTML]{1E1E1E} 0.69 ± 0.0744} & {\color[HTML]{1E1E1E} 0.70 ± 0.0584} & {\color[HTML]{1E1E1E} 0.69 ± 0.0610} \\
7 & Cylindrical & {\color[HTML]{1E1E1E} 0.75 ± 0.1912} & {\color[HTML]{1E1E1E} 0.58 ± 0.0859} & {\color[HTML]{1E1E1E} 0.64 ± 0.0908} & {\color[HTML]{1E1E1E} 0.84 ± 0.1570} & {\color[HTML]{1E1E1E} 0.62 ± 0.2558} & {\color[HTML]{1E1E1E} 0.67 ± 0.2046} \\ \hline
\end{tabular}
\label{tab:Imagenet60}
\end{table*}
\begin{table*}[ht]
\centering
\caption{Class-wise train and test performance for Depth Contrast pretrained and fine-tuned on \ac{3DPM} dataset in fully supervised setting}
\tiny
\begin{tabular}{cccccccc}
\hline
 &  & \multicolumn{6}{c}{Experiment No. 5} \\ \cline{3-8} 
S.No. & Classes & \multicolumn{3}{c}{Train} & \multicolumn{3}{c}{Test} \\ \cline{3-8} 
 &  & Precision & Recall & F1 & Precision & Recall & F1 \\ \hline
1 & Mixed 1 & {\color[HTML]{1E1E1E} 0.30 ± 0.0509} & {\color[HTML]{1E1E1E} 0.48 ± 0.1288} & {\color[HTML]{1E1E1E} 0.36 ± 0.0630} & {\color[HTML]{1E1E1E} 0.26 ± 0.0393} & {\color[HTML]{1E1E1E} 0.56 ± 0.2023} & {\color[HTML]{1E1E1E} 0.35 ± 0.0675} \\
2 & Agglomerated & {\color[HTML]{1E1E1E} 0.64 ± 0.0295} & {\color[HTML]{1E1E1E} 0.87 ± 0.0270} & {\color[HTML]{1E1E1E} 0.74 ± 0.0172} & {\color[HTML]{1E1E1E} 0.66 ± 0.0586} & {\color[HTML]{1E1E1E} 0.85 ± 0.0552} & {\color[HTML]{1E1E1E} 0.74 ± 0.0311} \\
3 & Mixed 2 & {\color[HTML]{1E1E1E} 0.70 ± 0.1256} & {\color[HTML]{1E1E1E} 0.32 ± 0.1026} & {\color[HTML]{1E1E1E} 0.43 ± 0.1033} & {\color[HTML]{1E1E1E} 0.63 ± 0.3049} & {\color[HTML]{1E1E1E} 0.20 ± 0.1086} & {\color[HTML]{1E1E1E} 0.29 ± 0.1546} \\
4 & Ore 1 & {\color[HTML]{1E1E1E} 0.93 ± 0.0226} & {\color[HTML]{1E1E1E} 0.71 ± 0.0198} & {\color[HTML]{1E1E1E} 0.80 ± 0.0193} & {\color[HTML]{1E1E1E} 0.90 ± 0.0330} & {\color[HTML]{1E1E1E} 0.73 ± 0.0380} & {\color[HTML]{1E1E1E} 0.81 ± 0.0179} \\
5 & Ore 2 & {\color[HTML]{1E1E1E} 0.88 ± 0.0555} & {\color[HTML]{1E1E1E} 0.72 ± 0.0391} & {\color[HTML]{1E1E1E} 0.79 ± 0.0278} & {\color[HTML]{1E1E1E} 0.90 ± 0.0361} & {\color[HTML]{1E1E1E} 0.74 ± 0.0404} & {\color[HTML]{1E1E1E} 0.81 ± 0.0250} \\
6 & Ore 3 & {\color[HTML]{1E1E1E} 0.68 ± 0.0577} & {\color[HTML]{1E1E1E} 0.78 ± 0.0307} & {\color[HTML]{1E1E1E} 0.73 ± 0.0333} & {\color[HTML]{1E1E1E} 0.73 ± 0.0365} & {\color[HTML]{1E1E1E} 0.74 ± 0.0510} & {\color[HTML]{1E1E1E} 0.73 ± 0.0193} \\
7 & Cylindrical & {\color[HTML]{1E1E1E} 0.96 ± 0.0359} & {\color[HTML]{1E1E1E} 0.45 ± 0.1523} & {\color[HTML]{1E1E1E} 0.60 ± 0.1421} & {\color[HTML]{1E1E1E} 0.89 ± 0.1535} & {\color[HTML]{1E1E1E} 0.56 ± 0.2940} & {\color[HTML]{1E1E1E} 0.62 ± 0.2551} \\ \hline
\end{tabular}
\label{tab:ssl80}
\end{table*}
\begin{table*}[ht]
\centering
\caption{Class-wise train and test performance for Depth Contrast pretrained and fine-tuned on \ac{3DPM} dataset in semi supervised setting using only 20\% of labelled data}
\tiny
\begin{tabular}{cccccccc}
\hline
 &  & \multicolumn{6}{c}{Experiment No. 6} \\ \cline{3-8} 
S.No. & Classes & \multicolumn{3}{c}{Train} & \multicolumn{3}{c}{Test} \\ \cline{3-8} 
 &  & Precision & Recall & F1 & Precision & Recall & F1 \\ \hline
1 & Mixed 1 & {\color[HTML]{1E1E1E} 0.35 ± 0.1273} & {\color[HTML]{1E1E1E} 0.51 ± 0.0632} & {\color[HTML]{1E1E1E} 0.41 ± 0.0763} & {\color[HTML]{1E1E1E} 0.30 ± 0.0816} & {\color[HTML]{1E1E1E} 0.44 ± 0.0785} & {\color[HTML]{1E1E1E} 0.34 ± 0.0340} \\
2 & Agglomerated & {\color[HTML]{1E1E1E} 0.56 ± 0.0657} & {\color[HTML]{1E1E1E} 0.90 ± 0.0470} & {\color[HTML]{1E1E1E} 0.69 ± 0.0469} & {\color[HTML]{1E1E1E} 0.52 ± 0.0757} & {\color[HTML]{1E1E1E} 0.85 ± 0.0548} & {\color[HTML]{1E1E1E} 0.64 ± 0.0575} \\
3 & Mixed 2 & {\color[HTML]{1E1E1E} 0.44 ± 0.5110} & {\color[HTML]{1E1E1E} 0.09 ± 0.0673} & {\color[HTML]{1E1E1E} 0.11 ± 0.0694} & {\color[HTML]{1E1E1E} 0.17 ± 0.3249} & {\color[HTML]{1E1E1E} 0.07 ± 0.0758} & {\color[HTML]{1E1E1E} 0.07 ± 0.0924} \\
4 & Ore 1 & {\color[HTML]{1E1E1E} 0.87 ± 0.0761} & {\color[HTML]{1E1E1E} 0.76 ± 0.0467} & {\color[HTML]{1E1E1E} 0.81 ± 0.0221} & {\color[HTML]{1E1E1E} 0.83 ± 0.0810} & {\color[HTML]{1E1E1E} 0.73 ± 0.0481} & {\color[HTML]{1E1E1E} 0.77 ± 0.0374} \\
5 & Ore 2 & {\color[HTML]{1E1E1E} 0.87 ± 0.0645} & {\color[HTML]{1E1E1E} 0.65 ± 0.1050} & {\color[HTML]{1E1E1E} 0.73 ± 0.0587} & {\color[HTML]{1E1E1E} 0.87 ± 0.0423} & {\color[HTML]{1E1E1E} 0.64 ± 0.1256} & {\color[HTML]{1E1E1E} 0.73 ± 0.0870} \\
6 & Ore 3 & {\color[HTML]{1E1E1E} 0.77 ± 0.1103} & {\color[HTML]{1E1E1E} 0.56 ± 0.1603} & {\color[HTML]{1E1E1E} 0.63 ± 0.1199} & {\color[HTML]{1E1E1E} 0.76 ± 0.1011} & {\color[HTML]{1E1E1E} 0.52 ± 0.1292} & {\color[HTML]{1E1E1E} 0.61 ± 0.1029} \\
7 & Cylindrical & {\color[HTML]{1E1E1E} 0.74 ± 0.4333} & {\color[HTML]{1E1E1E} 0.31 ± 0.2556} & {\color[HTML]{1E1E1E} 0.41 ± 0.2570} & {\color[HTML]{1E1E1E} 0.64 ± 0.4165} & {\color[HTML]{1E1E1E} 0.22 ± 0.2222} & {\color[HTML]{1E1E1E} 0.30 ± 0.2547} \\ \hline
\end{tabular}
\label{tab:ssl20}
\end{table*}
This section analyzes the results on downstream tasks of material classifications while comparing the effect of the proposed pretraining method. It also extends the analysis of the learning capabilities of models concerning labeled data quantity and different training strategies, including fine-tuning and linear evaluation.

Results for experiments 1, 2, and 3.1 mentioned in Table \ref{tab:exp} provides clear insights about input mode that combined representation of raw and reflectance image obtains higher performance on train and test set for fine-tuning on material classification. Commonly used Efficient-net b2 \cite{tan2020efficientnet} model with the combined representation of raw and reflectance image input achieves an F1 score of 0.71 on the test set whereas the model with only reflectance image remain lower F1 score of 0.69 and model with only raw images performs significantly poor. These experiments direct the current work to use the combined representation of raw and reflectance images for further experiments to evaluate different CNN architectures and the proposed self-supervised method.

Further, the explorations of DenseNet 121 \cite{huang2018densely} and ResNeXt (31x4d) \cite{xie2017aggregated} CNN architectures obtains F1 score of 0.67 and 0.72 in experiments 3.2 and 3.3 respectively. It shows that Efficient-net b2 \cite{tan2020efficientnet} in experiment 3.1 remains competitive concerning classification performance on the test set, lower standard deviation, and less prone to overfit on the train set. It is important to note that Efficient-net b2 \cite{tan2020efficientnet} being a lightweight model with lesser learning parameters, is a good fit for the 3DPM dataset. Analysis from experiments 1 to 3.3 indicates that the combined representation of raw and reflectance image with Efficient-net b2 \cite{tan2020efficientnet} architecture obtains benchmarked results and is a suitable candidate for further investigation on the proposed Depth Contrast method.
The Depth Contrast pretrained model, which is fine-tuned in a fully supervised setting in experiment 5, outperforms other models. It not only achieves the highest F1 score of 0.73 with minimal standard deviation but also depicts no overfitting with the same F1 score on the train set. 

More specifically, while comparing the performance of this model with ImageNet \cite{imagenet_cvpr09} pretrained model in the same fully supervised settings, it shows an improvement of 1\% on the test set. This improvement in the downstream material classification task showcases the leverage of unlabeled data used by the proposed Depth Contrast pretraining method to learn efficient representations without human annotations.

Interestingly, The Depth Contrast pretrained model, which is fine-tuned in a semi-supervised setting using only 20\% of labeled data in experiment 6, achieves a comparable F1 score of 0.65. The proposed method, Depth Contrast, clearly outperforms by 11\% on ImageNet \cite{imagenet_cvpr09} pretrained model in the same semi-supervised settings mentioned in experiment 4. This comparative study showcases the representation learning ability of the Depth Contrast method to yield performance generalization on unseen data even when trained on significantly reduced labeled data.

A similar performance is followed for linear evaluation tasks when pretrained CNN serves as a feature extractor, and only fully connected layers are trained. Table \ref{tab:exp_le} showcases a comparative analysis of the Depth Contrast pretrained model with the ImageNet pretrained model in fully and semi-supervised settings. Depth Contrast pretrained and linear evaluated model in experiment 9 obtains F1 score 0.72, which is 3\% higher than ImageNet pretrained model in experiment 7 for same fully supervised setting. Similarly, Depth Contrast pretrained model from experiment 10 obtains F1 score of 0.64, which is 12\% higher than the ImageNet model in the same semi-supervised setting. While linear evaluation clearly shows that ImageNet pretrained model suffers high overfitting in limited labeled data availability, whereas Depth Contrast leverages domain-specific unlabeled data and learns efficient representations for performance generalization.

Detailed class-wise results are mentioned in Table \ref{tab:Imagenet60} and Table \ref{tab:ssl80} for ImageNet and Depth Contrast pretrained model respectively in fully supervised setting. It is observed that material categories compounded by multiple materials, Mixed1 and Mixed2, perform poorly since their texture characteristics are not uniform, and the sample count is also low in the 3DPM dataset. A similar trend is followed in Depth Contrast pretrained model in the semi-supervised setting mentioned in Table \ref{tab:ssl20}. However, the recall and precision of other classes are better for the Depth Contrast pretrained model. Further, current work shares the following conceptual findings.
\begin{figure}[h]
    \centering
    \includegraphics[width =0.5\linewidth]{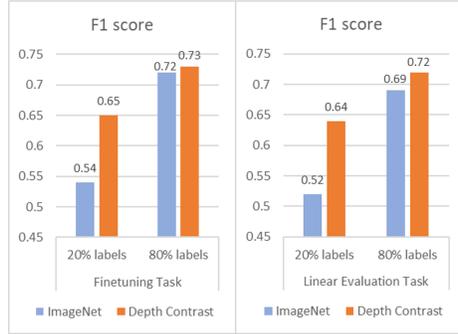}
    \caption{Performance comparison of models pretrained on ImageNet and Depth Contrast Efficient-net in semi and fully supervised settings}
    \label{fig:f1}
\end{figure}
\subsection{Raw image of depth maps scales representations and improves learning}
Depth maps of the material on conveyor belts are directly associated with the shape and size of the material; thus, depth details are valuable information addition for learning representation. However, alone depth maps are not sufficient to define the material. It provides the reason that the combined representation of raw (depth details) and reflectance images yield optimal performance. This fact also forms the backbone of the Depth Contrast pretraining method for view pair modeling. 
\subsection{CNN with reduced parameters performs better}
The architecture of Efficient-net b2 \cite{tan2020efficientnet} offers adaptive properties for width, depth, and resolution of convolutional layers based on input image size. It seems that the model adapts conveniently based on dataset size and texture details, making it a suitable CNN architecture even with fewer learning parameters.
\subsection{Self-supervised method Depth Contrast significantly improves downstream task performance and reduces human annotations}
As Depth Contrast constructs input pair view using reflectance image and depth map of the raw image, which allows learning representation comprehensively using both types of details textures from reflectance view and depth maps. During contrastive mechanism, it learns correlation between texture and depth of material, whereas negative pairs allow learning dissimilarity among texture and depth details across material categories. Performance on downstream tasks supports the fact about learning representations in the self-supervised manner as the method outperforms in fully and semi-supervised labeled data settings. Specifically, in limited labeled data scenarios, performance remains statistically significant (p \textless 0.05).
\subsection{Self-supervised method Depth contrast encourages performance generalization}
Comparative analysis of fine-tuning and linear evaluation task in Figure \ref{fig:f1} shows that gap in classification performance between ImageNet pretrained and Depth Contrast pretrained model becomes significant when moving from fine-tuning to linear evaluation. It indicates the rich representation learning capabilities of the model due to Depth Contrast that when CNN model is used as only feature extractor, then also a performance on unseen data remain competent whereas the ImageNet pretrained model starts suffering in performance generalization.  
It also shows the potential of the self-supervised methods in the visual domain beyond the natural visual concepts that are challenging to understand and annotate by humans.

\section{Conclusion and Future Scope}\label{Conclusion and Future Scope}
Current work explores the computer vision methods and deep \ac{CNNs} on mining material classification on \ac{3DPM} sensor images. It proposes the self-supervised learning method Depth Contrast to improve the representation learning by leveraging sensor-generated data without human annotation. This work contributes the following:
\begin{itemize}
    \item Obtains benchmark results of F1 score 0.72 on 3DPM dataset for mining material classification employing ImageNet \cite{imagenet_cvpr09} transfer learning in a fully supervised setting
    \item Contributes to a novel self-supervised pretraining method, Depth Contrast, which improves performance for mining material classification by obtaining an F1 score of 0.73 in a fully supervised setting. It obtains F1 score 0.65 semi-supervised setting which shows improvement of 11\% over ImageNet \cite{imagenet_cvpr09} transfer learning. The proposed method depicts consistent performance improvement on fine-tuning and linear evaluation. It shows the applicability of self-supervised learning beyond the natural visual concept domain.
    \item The Depth Contrast method showcases the utilization and leverage of sensor-generated visual data to learn representations, significantly reducing the need for human supervision, thus making industrial AI automation more robust, cost-effective, and serving as lifelong learning.  The proposed method's main idea is based on multiple input views from different sensors, which has potential possibilities to apply the method in different automation scenarios where views are in terms of video frames, visual captures from different angles, environment, etc.
\end{itemize}
The potential of self-supervised methods in the work indicates the well-versed possibilities to utilize the machine-generated unlabeled data to reduce the need for human supervision beyond natural visual concepts, reducing the cost and increasing efficiency. The proposed method is adaptable in all the domains where depth information is significant. Future work is subtle to investigate further the field of contrastive joint embedding and other self-supervised methods based on knowledge distillations and bootstrapping. Exploring more network architectures, including transformers, is also essential to elevate supervised performance with increased data.

\clearpage
%
%
\bibliographystyle{splncs04}
\bibliography{mybibliography}
\end{document}